\documentclass[10pt,twocolumn,letterpaper]{article}

\usepackage[pagenumbers]{cvpr} 

\usepackage{url}
\usepackage{multirow}
\usepackage{floatrow}
\usepackage{tcolorbox}
\usepackage{threeparttable}
\usepackage{color}
\usepackage{soul}
\usepackage{graphicx}
\usepackage{amsmath}
\usepackage{amssymb}
\usepackage{booktabs}
\usepackage{tabularx}
\usepackage{colortbl}
\usepackage{float}
\usepackage{wrapfig}
\usepackage{ragged2e}
\definecolor{dt}{gray}{0.6}

\def\eg{\emph{e.g., }}

\def\name{VideoPerceiver}

\def\fig{Fig. }
\def\tab{Tab. }

\usepackage{xcolor}

\definecolor{cvprblue}{rgb}{0.21,0.49,0.74}
\usepackage[pagebackref,breaklinks,colorlinks,allcolors=cvprblue]{hyperref}

\def\paperID{19076} 
\def\confName{CVPR}
\def\confYear{2026}

\title{\name: Enhancing Fine-Grained Temporal Perception in Video Multimodal Large Language Models}

\author{\textbf{Fufangchen Zhao}\textsuperscript{\rm1},
\textbf{Liao Zhang}\textsuperscript{\rm2},
\textbf{Daiqi Shi}\textsuperscript{\rm2},
\textbf{Yuanjun Gao}\textsuperscript{\rm2},
\textbf{Chen Ye}\textsuperscript{\rm2},
\textbf{Yang Cai}\textsuperscript{\rm2},
\\
\textbf{Jian Gao}\textsuperscript{\rm1},
\textbf{Danfeng Yan}$^*$\textsuperscript{\rm1}
\\
\textsuperscript{\rm1}State Key Laboratory of Networking and Switching Technology, BUPT\\
\textsuperscript{\rm2}Independent Researcher$^\dagger$\\ 
{\tt\small \{zhaofufangchen, gaojian, yandf\}@bupt.edu.cn,}
}

\begin{document}
\maketitle

\def\thefootnote{$\dagger$}\footnotetext{Due to the requirements of the author's institution, institutional information will not be displayed}
\def\thefootnote{*}\footnotetext{Corresponding author}

\begin{abstract}

We propose \name, a novel video multimodal large language model (VMLLM) that enhances fine-grained perception in video understanding, addressing VMLLMs’ limited ability to reason about brief actions in short clips or rare transient events in long videos. \name adopts a two-stage training framework. During supervised fine-tuning (SFT), we construct “key-information-missing” videos by extracting event-action keywords from captions, identifying corresponding key frames, and replacing them with adjacent frames. We jointly encode original and modified video tokens with text tokens, aligning intermediate visual representations with keywords via an auxiliary contrastive loss to enhance sensitivity to fine-grained motion cues. In reinforcement learning (RL), both video variants are fed into the model to generate descriptions, and a novel relative reward ensures responses from complete videos outperform those from degraded inputs, explicitly training the model to recover temporally precise action details. We also curate a dataset of 80,000 videos with fine-grained actions and transient events. Experiments show \name substantially outperforms state-of-the-art VMLLMs on fine-grained action understanding and rare event captioning benchmarks, while maintaining strong performance on standard tasks. By prioritizing task-relevant visual features, our work redefines video-language model training for fine-grained perception.
\end{abstract}    
\section{Introduction}
\label{sec:intro}
\begin{figure}[t]
    \centering
    \includegraphics[width=0.98\linewidth]{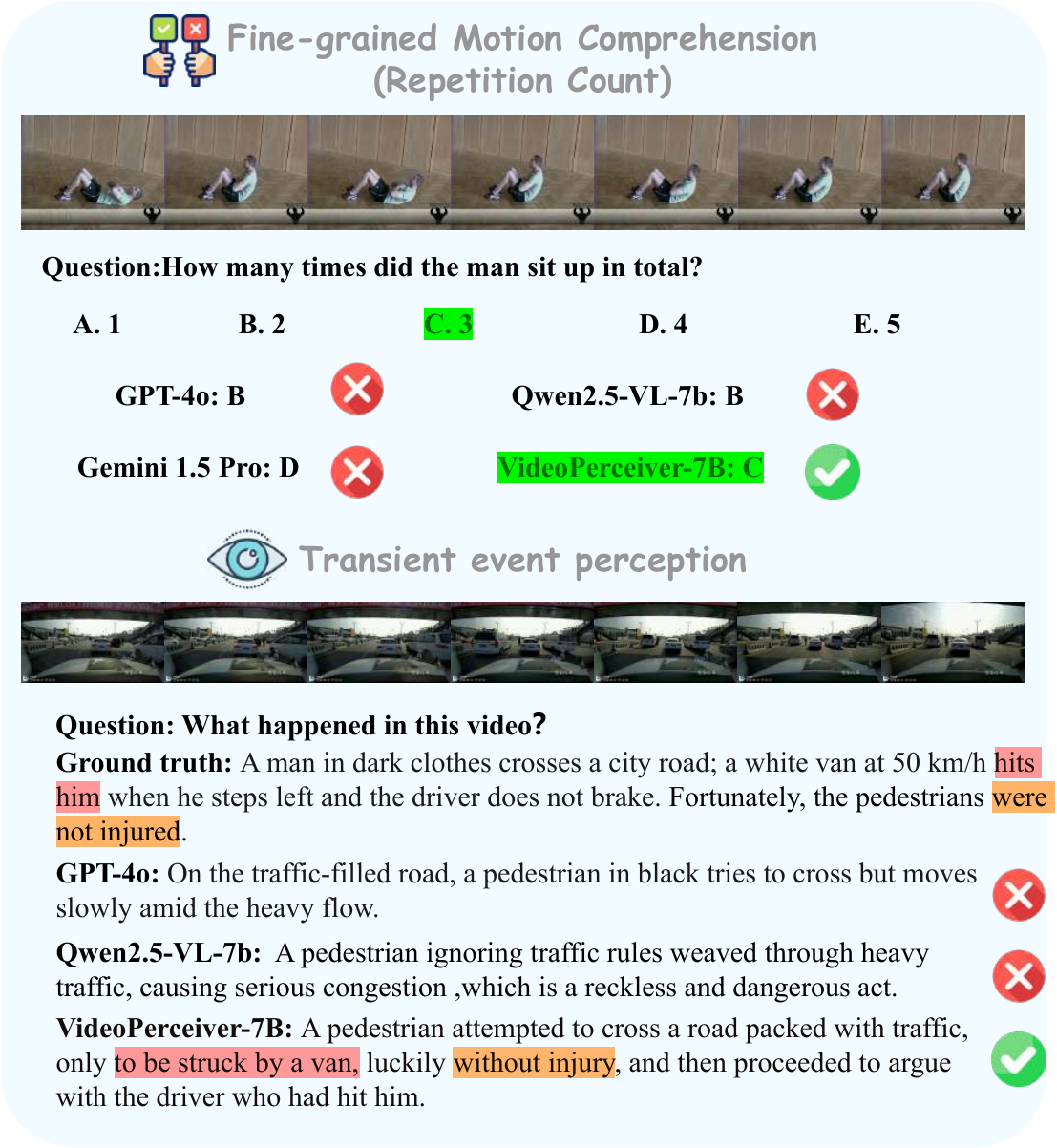}
    \caption{The illustration of the proposed \name.}
    \label{examples}
\end{figure}

The remarkable success of multimudal large language models (MLLM) cexemplified by architectures such as CLIP \cite{li2023blip}, Flamingo \cite{alayrac2022flamingo}, LLaVA \cite{zhang2024llava} and Qwen-vl series \cite{wang2024qwen2,Qwen2.5-VL} has catalyzed a paradigm shift in multimodal understanding, enabling machines to reason about visual information with unprecedented linguistic fluency.  Recently, this frontier has expanded into temporal domain, giving rise to video-multimodal large language models (VMLLM) that aim bridge dynamic visual content with natural language \cite{maaz2023video,li2023videochat,li2024mvbench,maaz2024videogpt+,li2024aria,liu2024st,liu2024oryx,team2024gemini}. For example, The VideoChat series \cite{li2023videochat, li2024mvbench} encodes video frames using a visual encoder and transforms them into fixed-length visual tokens via Q-Former \cite{li2023blip}, these visual tokens share the same dimensionality as textual tokens and can be directly input to large language models such as Mistral \cite{jiang2023mistral7b}, GLM \cite{glm2024chatglm}, and Qwen \cite{qwen2,qwen2.5,qwen3}. LLaVA-Video \cite{zhang2024video} and PLLaVA \cite{xu2024pllava} extend static vision-language pretraining by further training multimodal large language models (MLLMs) on dynamic video for video understanding. Moreover, Qwen2.5-VL \cite{Qwen2.5-VL} and LLaVA-OneVision \cite{li2024llava} reduce the training cost of MLLMs by jointly training on both images and videos. 

Recently, researchers have observed that integrating reinforcement learning (RL) reward mechanisms into MLLMs can enhance visual understanding by strengthening the models' reasoning capabilities \cite{shen2025vlm,jiang2025vlm,wang2025vl}. This approach has since been extended to video understanding tasks. Specifically, Video-R1 \cite{feng2025video} incorporates a temporal reward mechanism based on GRPO \cite{shao2024deepseekmath}, VideoChat-R1 \cite{li2025videochat} integrates reinforcement learning directly during fine-tuning, and VITAL \cite{zhang2025thinking} applies RL to improve the tool-use capability of video VMLLMs. These advances have collectively contributed to the progress of VMLLMs.

Despite rapid progress, a critical gap persists: current video VLMs remain fundamentally limited in their capacity to perceive and reason about fine-grained temporal motion events, whether brief actions in short clips (\eg flicking a switch or a shift in facial expression) or rare transient events embedded in long videos (\eg a traffic accident in surveillance footage) as show in \fig \ref{examples}.

This limitation arises from prevailing temporal sampling strategies \cite{Qwen2.5-VL,wang2024qwen2,zhu2025internvl3,chen2024expanding,chen2024internvl} and text-centric reward designs in reinforcement learning \cite{shao2024deepseekmath}. First, existing VMLLMs employ uniform frame sampling \cite{Qwen2.5-VL} or fixed frame windows \cite{chen2024internvl}, which either discard critical fine-grained events through aggressive downsampling or lack the temporal span required to capture long-range contextual cues for disambiguation. Second, Transformer-based visual encoders typically process all sampled frames uniformly, without adaptively attending to those containing salient visual cues, thereby prioritizing holistic event-level understanding over localized discriminative details \cite{dosovitskiy2020image, liu2021swin}. Moreover, current reinforcement learning \cite{shao2024deepseekmath} continue to define rewards solely based on properties of the generated text (\eg format, length, and quality). Although recent methods \cite{feng2025video,li2025videochat,zhang2025thinking} have incorporated visual awareness into reward shaping, they remain insufficient in mitigating the above limitations. As a result, existing video VMLLMs frequently fail to distinguish between temporally concentrated yet fine-grained actions or entirely miss sparse but semantically pivotal moments.

To address this bottleneck, we introduce \name, a novel MLLM specifically designed to perceive subtle actions and short-duration events across videos of variable length. Our approach leverages a two-stage training framework. First, during supervised fine-tuning (SFT), we construct a “key-information-absent” video by extracting event-action keywords from captions, identifying corresponding keyframes, and replacing them with neighboring frames. The original and degraded video tokens along with text tokens are jointly encoded, and intermediate visual representations are aligned with event-action keywords via an auxiliary contrastive loss \cite{he2020momentum, radford2021learning} to sharpen sensitivity to fine-grained motion cues. Second, in reinforcement learning (RL), both video variants are sequentially input to elicit two descriptions, a novel relative reward enforces that the second response (based on the full video) must be superior to the first (based on the degraded input), and this signal is integrated into GRPO \cite{shao2024deepseekmath}. This design explicitly trains the model to recover and articulate temporally precise action details, significantly enhancing its capacity for fine-grained video understanding.

To further enhance \name’s performance on fine-grained action and transient event perception tasks, we extract and curate 80K videos containing fine-grained actions and short-duration events from open-source, publicly annotated datasets including HMDB51 \cite{kuehne2011hmdb}, CelebV-HQ \cite{zhu2022celebv}, and MM-AU \cite{fang2024abductive} and generate detailed captions along with question-answer pairs targeting fine-grained action and transient event perception using GPT-4o\cite{openai2024gpt4o} and GPT-4 \cite{achiam2023gpt}. Moreover, for videos containing short-duration events, we precisely localize the event-bearing frames and augment the input sequence with non-event frames to reduce the temporal density of the target event within the video.


We demonstrate that \name~ significantly outperforms state-of-the-art VMLLMs on benchmarks for fine-grained action understanding (MotionBench \cite{hong2025motionbench}) and rare event captioning (VRU-Accident \cite{kim2025vru}) tasks where conventional approaches exhibit consistent underperformance. Moreover, when evaluated beyond the scope of fine-grained actions and transient events, \name~ maintains strong performance on standard VMLLMs benchmarks \cite{li2024mvbench,fu2025video}. More broadly, our work redefines the design philosophy of video language models that we propose a training paradigm that prioritizes task-relevant visual characteristics, wherein the model actively detects, models, and reasons about fine-grained or transient visual semantics, rather than treating video as a sequence of static frames subjected to averaging or uniform encoding. This approach not only enhances performance on established tasks but also enables robust capabilities in domains requiring fine-grained motion awareness and high temporal acuity (\eg detailed action recognition, interpretation of transient accident events in surveillance footage) offering a practical and principled direction for future research in video-language understanding.

In summary, our main contributions are list as follows:
\begin{itemize}

    \item We propose \name, which, to the best of our knowledge, is the first video multimodal large language model (VMLLM) explicitly designed for perceiving fine-grained temporal actions and transient events.
    \item We introduce a key-information-absent video construction strategy and employ contrastive learning to enhance the model’s sensitivity during SFT Stage to subtle motion cues and critical transient events.
    \item We devise a comparable reinforcement learning reward mechanism that compares the description quality of outputs generated from the original video and its degraded counterpart, thereby guiding the model to accurately reconstruct fine-grained actions and transient event details.
    \item We curate a large-scale, high-quality dataset comprising 80K videos featuring fine-grained actions and transient events for both SFT and RL training, and further increase temporal localization difficulty by sparsifying transient events through non-event frame augmentation.
\end{itemize}

\section{Related Work}
\label{sec:Related work}

\subsection{Video Multimodal Large Language Models}

Current VMLLMs grapple with two core challenges: (a) managing the surge in visual tokens from long videos, and (b) retaining fine-grained temporal dynamics during feature compression. Existing solutions primarily differ in projector design. MLP-based projectors (e.g., Video-ChatGPT \cite{maaz2023video}, PLLaVA \cite{xu2024pllava}) offer simplicity but process frames independently, neglecting temporal dependencies. In contrast, Q-Former-based approaches \cite{li2023blip} use learnable queries to produce compact, length-invariant representations, naturally supporting variable-length inputs. Notable variants include Vista-LLaMA \cite{ma2023vista} with recurrent temporal attention, VideoChat \cite{li2023videochat,li2024mvbench} with dedicated temporal alignment queries, and ST-LLM \cite{liu2024st}, which reuses an image-pretrained Q-Former without extra parameters.

Most VMLLMs still employ image-centric backbones such as CLIP ViT \cite{radford2021learning} or Eva-CLIP ViT \cite{sun2023eva}, which lack inherent temporal modeling. Recent works mitigate this by adopting video-native encoders. For instance, VideoChat2 \cite{li2024mvbench} uses UMT-L \cite{li2023unmasked}, and Video-LLaVA \cite{lin2023video} leverages LanguageBind’s video encoder \cite{zhu2023languagebind}. Nevertheless, these methods often over-compress spatiotemporal features and treat all frames uniformly, thereby losing critical local details and failing to emphasize task-relevant key moments.

\subsection{Transient Fine-grained Motion and Event Video Understanding}


Early approaches to fine-grained action perception relied on sparse annotations or structured video captions \cite{soomro2012ucf101,kuehne2011hmdb,caba2015activitynet,ju2024miradata,fan2025instancecap,wu2025any2caption}. However, these methods typically lack explicit semantic representations of actions. Recent benchmarks such as MotionBench \cite{hong2025motionbench} and FAVOR-Bench \cite{tu2025favor} address this gap by incorporating semantic labels for fine-grained actions. Datasets like CelebV-HQ \cite{zhu2022celebv} and MotionVid-QA \cite{du2025motionsight}, owing to their large scale and fine-grained textual annotations, exhibit strong utility as training resources. On the modeling side, FaVChat \cite{zhao2025favchat} improves facial action understanding through multi-granularity feature integration, while Human-Omni \cite{zhao2025humanomni} employs multiple encoders to achieve holistic representation of the primary subject. Despite these advances, existing methods remain limited in capturing transient, fine-grained actions, a capability that \name~ is specifically designed to enhance.

Transient event perception seeks to accurately identify critical short-duration events (typically 1–2 seconds) within long videos (minutes to hours), a problem commonly framed as temporal grounding (TG). To enable precise event localization, VTimeLLM \cite{huang2024vtimellm} fixes the video frame sequence and encodes frame indices as explicit temporal coordinates. TimeChat \cite{ren2024timechat} and TimeSuite \cite{zeng2024timesuite} integrate timestamp awareness into their visual encoders, while Qwen2.5-VL \cite{Qwen2.5-VL} adopts MRoPE for temporal modeling. Similarly, TimeMaker \cite{chen2024timemarker} and VideoLLaMA-3 \cite{zhang2025videollama} prefix each video frame with a fixed timestamp token. Although these approaches incorporate temporal cues and achieve moderate success in localizing salient events, they remain ineffective for extremely sparse events such as traffic accidents in surveillance footage that last only 1–2 seconds and represent less than 1\%, often below 0.1\%, of the total video duration. This limitation underscores the need for a more sensitive temporal perception mechanism, motivating the design of our \name.

\subsection{Reinforcement Learning for VMLLMs}

Following the remarkable success of MLLMs \cite{deng2025boosting,deng2025openvlthinker,liu2025seg,liu2025visual,peng2025lmm,yang2025r1,zhan2025vision,zhang2025r1,zhou2025r1}, reinforcement learning has increasingly been extended to video understanding. Early approaches such as TimeZero \cite{wang2025timezero} and R1-Omni \cite{zhao2025r1} apply GRPO \cite{shao2024deepseekmath} to video tasks involving temporal distribution modeling and sentiment analysis. Video-R1 \cite{feng2025video} enhances spatial understanding in videos by incorporating spatiotemporal ordering into GRPO. Building on this, STAR-R1 \cite{qistar} introduces a tailored spatiotemporal reward mechanism to further improve the model’s reasoning capability in dynamic, long-duration scenes. In the proposed \name, we incorporate a comparative reasoning framework into reinforcement learning and devise a reward mechanism based on relative quality assessment between generated outputs, yielding strong performance on tasks involving transient, fine-grained actions and events.

\begin{figure*}[t]
    \centering
    \includegraphics[width=\linewidth]{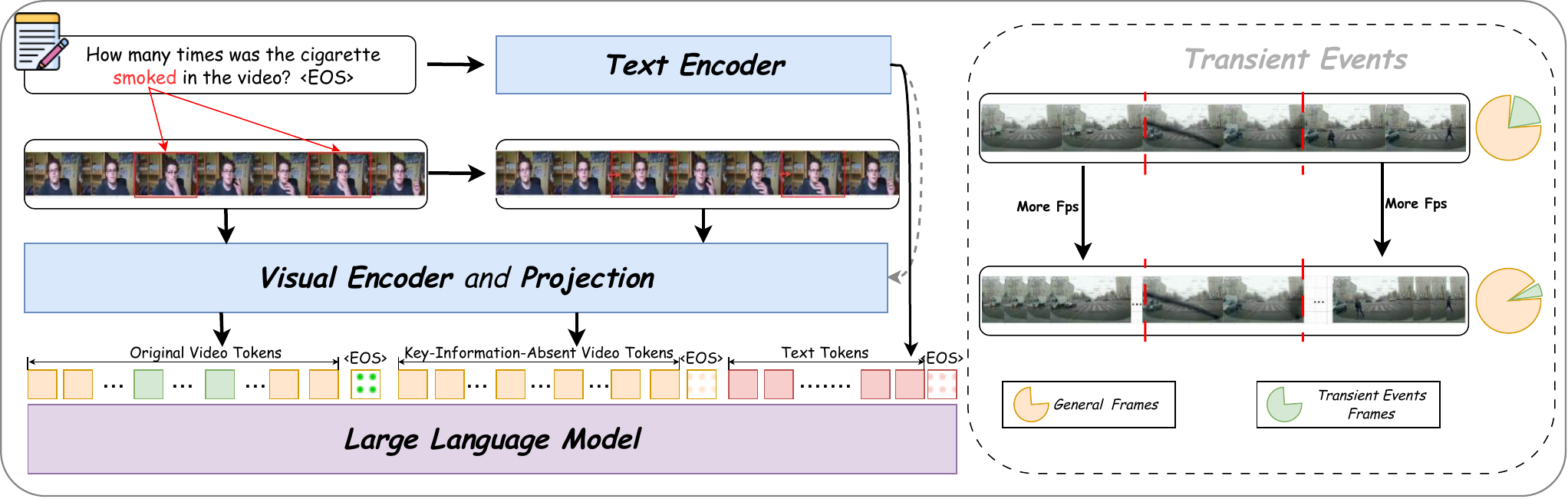}
    \caption{Key-information-absent video construction (left) and special frame sampling strategy for opportunity transient events (right).}
    \label{videopeceiver_input}
\end{figure*}

\section{\name}

This section presents a detailed overview of the proposed \name, which incorporates three key innovations: (i) the key-information-absent video construction strategy, (ii) a supervised fine-tuning (SFT) approach that leverages contrastive learning on intermediate model representations, and (iii) a reinforcement learning (RL) framework centered on comparative reward modeling. We now describe each of these components in detail.
\subsection{Key-Information-Absent Video Construction}
\label{key-information-absent video construction}

The Key-Information-Absent Video Construction module constitutes the first core innovation of the \name~ framework, as depicted in \fig \ref{videopeceiver_input}. Given a video $V=\{v_i\}_{i=1}^n$ paired with a textual prompt $T$ (\eg a question or caption), our goal is to synthesize a modified video $\tilde{V}$ that lacks semantically critical transient events while preserving overall visual coherence.

We begin by leveraging a LLM (\eg GPT-4 \cite{achiam2023gpt}) to extract a set of fine-grained keywords $\mathcal{K}=\left\{k_i\right\}_{i=1}^N$ that describe transient actions or events implied by $T$. Concurrently, we employ the vision-language alignment capability of BLIP-2 \cite{li2023blip} to compute frame-wise relevance scores between $\mathcal{K}$ and each frame $v_i$. we obtain its embedding $f_i = \mathrm{BLIP}(v_i)$ and the pooled keyword embedding ${k}=\mathrm{mean}(\{\mathrm{BLIP}(k_i)_{i=1}^N\})$, then compute the cosine similarity:
\begin{equation}
s_i=\frac{f_i^{\top} k}{\left\|f_i\right\|\|k\|}
\end{equation}

Key-information frames are identified as those at local maxima of the similarity sequence $\{s_i\}_{i=1}^n$ i.e., 
\begin{equation}
\mathcal{T}_{\text {key }}=\left\{t \mid s_t>s_{t-1} \text { and } s_t>s_{t+1}\right\},
\end{equation}
with boundary conditions handled appropriately.

To construct the key-information-absent video $\hat{V}$ we replace each key frame $v_t$ ($\mathrm{for}~ t \in \mathcal{T}_{\text {key }}$) we replace each key frame with its immediate predecessor $v_{t-1}$ (or $v_t$ itself if $t=1$), yielding $\hat{v}_t=v_{t-1} ~\mathrm{for}~ t\in \mathcal{T}_{\text {key }}$, and $\tilde{v}_t=v_t$ otherwise.

Both $V$ and $\hat{V}$ are processed by a shared visual encoder, producing spatiotemporal representations that are temporally pooled and projected into token sequences $H_{v}$ and $\hat{H}_{v}$, respectively. The text $T$ is encoded via a text encoder into token sequence $H_t$. The final input to the multimodal transformer is constructed by concatenating these components with dedicated end-of-sequence tokens:
\begin{equation}
\small
\mathbf{H}=\left[H_v;<\mathrm{EOS}>_{H_{v}};\hat{H}_v;<\mathrm{EOS}>_{\hat{H}_{v}};H_t;<\mathrm{EOS}>_{H_{t}};\right],
\end{equation}
enabling the model to learn discriminative representations by contrasting full and key-absent visual contexts.

Notably, standard transient-event datasets such as MM-AU \cite{fang2024abductive} exhibit an unnaturally high density of key events, often exceeding 25\% of total frames, which biases models toward trivial solutions. To mitigate this, we apply denser temporal sampling (e.g., 8fps) exclusively to non-key segments during preprocessing, thereby reducing the effective key-event ratio to $\leq$5\% and better simulating real-world scenarios where critical moments are sparse.

\subsection{Intermediate Layer Contrastive Learning} 
\label{sft}
\begin{figure}[h]
    \centering
    \includegraphics[width=\linewidth]{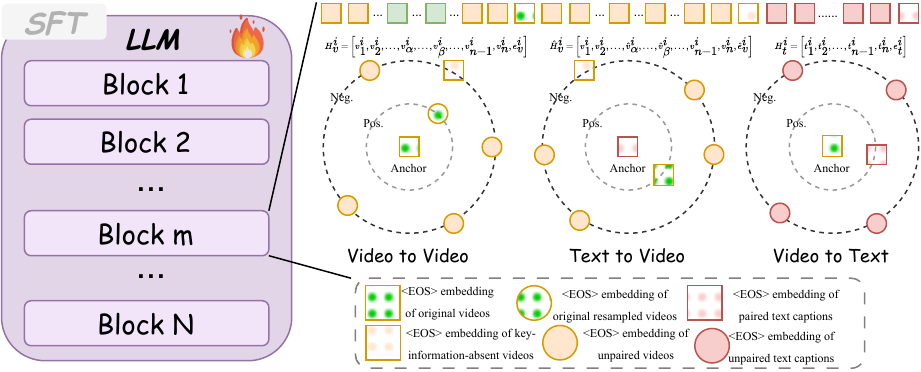}
    \caption{Schematic diagram of comparative learning in the intermediate layer.}
    \label{videopeceiver_sft}
\end{figure}

The second core innovation of \name~ is Intermediate Layer Contrastive Learning, which enhances the model’s sensitivity to transient events during supervised fine-tuning (SFT). As illustrated in \fig \ref{videopeceiver_sft}, we introduce three complementary contrastive objectives at a intermediate layer of LLM.

Based on Section \ref{key-information-absent video construction}, we first obtain the outputs $H_v^i=\left[v_1^i, v_2^i, \ldots, v_{\alpha}^i,\ldots,v_{\beta}^i,\ldots,v_{n-1}^i,v_n^i,e_v^i\right]$, $\hat{H}_v^i=\left[v_1^i, v_2^i, \ldots, \hat{v}_{\alpha}^i,\ldots,\hat{v}_{\beta}^i,\ldots,v_{n-1}^i,v_n^i,\hat{e}_v^i\right]$, and $H_t^i=\left[t_1^i, t_2^i, \ldots, t_{n-1}^i,t_n^i,e_t^i\right]$ from the  $m$-th LLM block, where $v_{\alpha}^i$ and $v_{\beta}^i$ denote video tokens that contain key information. Furthermore, the $<EOS>$ token $e_v^i$, $\hat{e}_v^i$ and $e_t^i$ as global representations capturing the complete semantic content of $H_v^i$, $\hat{H}_v^i$ and $H_t^i$, respectively. These $<EOS>$ tokens are subsequently employed in contrastive learning.

For video-to-video contrast, inspired by SimCSE \cite{gao2021simcse}, we generate a second view of the original video by re-encoding its visual tokens under a different stochastic dropout mask, yielding a perturbed embedding $\tilde{e}_v^i$, We then apply an InfoNCE-style contrastive loss:

\begin{equation}
\footnotesize
\begin{aligned}
& \mathcal{L}_{C L}^{v2v}= \\
& -\sum_{i=1: B+1} \frac{1}{B+1} \log \left[\frac{f\left(e_v^i, \tilde{e}_v^i\right)}{f\left(e_v^i, \hat{e}_v^i\right)+f\left(e_v^i, \tilde{e}_v^i\right)+\sum_{k \neq i} f\left(\hat{e}_v^i, \hat{e}_t^k\right)}\right]
\end{aligned}
\end{equation}

where $f\left(e_v^i, e_v^i\right)$ measures the distance between $e_v^i$ and $e_v^i$ in a semantic space. and $B$ denotes a training batch.

For cross-modal alignment, we define two additional losses. In text-to-video contrast, the text embedding $e_t^i$ serves as the anchor, with $e_v^i$ as the positive and $\mathcal{N}_{\text {t} 2 \mathrm{v}}=\left\{\hat{e}_v^i\right\} \cup\left\{{e}_{v}^k\right\}_{k \neq i}$ as negatives:
\begin{equation}
\footnotesize
    \mathcal{L}^{\mathrm{t2v}}_{CL}=-\sum^{\mathcal{N}_{\text{t} 2 \mathrm{v}}}\frac{1}{\mathcal{N}_{\text{t} 2 \mathrm{v}}}\mathrm{log}\left[\frac{f(e_t^i,e_v^i)}{f(e_t^i,e_v^i) + \sum_{e^{-} \in \mathcal{N}_{\text{t} 2 \mathrm{v}}}f(e_t^i,e^{-})}\right]
\end{equation}

Conversely, in video-to-text contrast, $e_v^i$ is the anchor, $e_t^i$ is the positive, and negatives are unpaired text embeddings $\mathcal{N}_{\mathrm{v} 2 \mathrm{t}}=\left\{\mathbf{e}_{\mathrm{text}}^j\right\}_{j \neq i}$:

\begin{equation}
\footnotesize
    \mathcal{L}^{\mathrm{v2t}}_{CL}=-\sum^{\mathcal{N}_{\text{v} 2 \mathrm{t}}}\frac{1}{\mathcal{N}_{\text{v} 2 \mathrm{t}}}\mathrm{log}\left[\frac{f(e_v^i,e_t^i)}{f(e_v^i,e_t^i) + \sum_{e^{-} \in \mathcal{N}_{\text{v} 2 \mathrm{t}}}f(e_v^i,e^{-})}\right]
\end{equation}

The video-to-video objective sharpens intra-video discriminability by emphasizing features that persist despite perturbation (typically those tied to salient transient events). The cross-modal objectives enforce tight coupling between textual keywords and their visual instantiations, ensuring that removal of key frames (as in $\hat{e}_v^i$) egrades alignment. Together, these losses guide the model to attend to fine-grained, temporally localized semantics rather than relying on coarse or redundant visual cues.

Motivated by VIRAL \cite{yoon2025visual}, we restrict contrastive supervision to the 16th LLM block which balances semantic abstraction and detail preservation avoiding per-layer application to maintain training efficiency.The total optimization object which jointly optimized with the primary SFT objective can be defined as follow:
\begin{equation}
\mathcal{O}=\arg \min \mathcal{L}_{SFT}+\left(\mathcal{L}_{C L}^{v2v}+\mathcal{L}_{C L}^{t2v} + \mathcal{L}_{CL}^{v2t}\right) / 3
\end{equation}

\subsection{GRPO with Quality comparator}
\label{rl}

\begin{figure*}[h]
    \centering
    \includegraphics[width=0.98\linewidth]{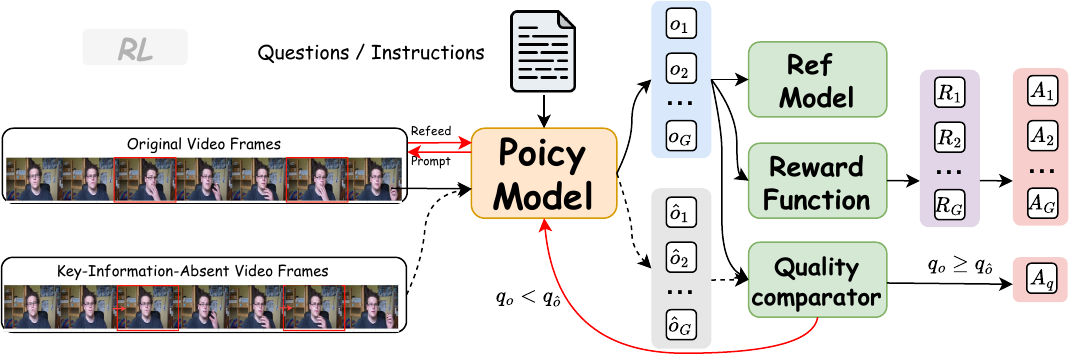}
    \caption{Schematic diagram of Comparative GRPO.}
    \label{videopeceiver_rl}
\end{figure*}

To further enhance temporal sensitivity during reinforcement learning, we propose Comparative GRPO, an extension of Group Relative Policy Optimization (GRPO \cite{shao2024deepseekmath}) that introduces a quality comparator to explicitly reward responses that depend on critical visual content. As illustrated in \fig \ref{videopeceiver_rl}, processes both the original video $V$ and its key-information-absent counterpart $\hat{V}$ with the same textual prompt $q$.

For each input, the policy model $\mathcal{\pi}_\theta$ samples $G$ responses: $\left\{o_i\right\}_{i=1}^G \sim \pi_\theta(\cdot \mid V, q)$ and $\left\{\hat{o}_i\right\}_{i=1}^G \sim \pi_\theta(\cdot \mid \hat{V}, q)$.  In parallel with the standard GRPO objective, which computes a base reward $R_{base}(o)$ based on surface attributes (\eg length, formatting, fluency), our quality comparator produces a comparative reward $R_{comp}$ defined as:
\begin{equation}
R_{\text {comp }}=\frac{1}{G} \sum_{i=1}^G\left[\mathcal{M}\left(o_i, o^{\text {ref }}\right)-\mathcal{M}\left(\hat{o}_i, o^{\text {ref }}\right)\right],
\end{equation}
where $o^{\text{ref}}$ is a ground-truth reference answer, and $\mathcal{M}(\cdot, \cdot)$ is a task-adaptive quality metric: For closed-form questions (\eg “How many times was the cigarette smoked?”), $\mathcal{M}=\mathbb{I}\left[y=y^{\mathrm{ref}}\right]$. For open-ended questions, $\mathcal{M}$ is the entailment probability from a pretrained NLI model, measuring whether $o$ is entailed by $o^{\text{ref}}$.

The comparative advantage is then estimated as:
\begin{equation}
A_{\text{comp}}=R_{\text{comp}}-b_{\text{comp}},
\end{equation}
where $b_{\text{comp}}$ is a baseline (\eg exponential moving average of past $R_{\mathrm{comp}}$).

Crucially, if $R_{\mathrm{comp}} \leq 0$, which indicates the model fails to leverage key visual information, we activate a refeed mechanism: the system appends an instruction such as “\textit{Re-answer with attention to fine-grained visual details}” to $q$, and resamples responses from $\mathrm{\pi}_\theta$ for $(V,q)$ until $R_{\mathrm{comp}} > 0$ or a maximum retry limit is reached. This ensures the policy gradient is only updated when the model demonstrates discriminative perception.

The total advantage combines both signals:

\begin{equation}
    A_{\mathrm{total}} = A_{\mathrm{base}} + \lambda A_{\mathrm{comp}},
\end{equation}
where $\lambda$ balances base and comparative rewards. The policy is updated via GRPO’s group-wise objective using $A_{\mathrm{total}}$, explicitly incentivizing responses that degrade gracefully when key visual content is removed, thereby internalizing fine-grained temporal awareness.

\subsection{data Craft}

To support the fine-grained temporal perception capabilities of \name, we construct a specialized dataset, denoted \name-80K, comprising 80k video clips that exhibit fine-grained actions or transient events (defined as visual changes lasting $\leq$ 1 second or spanning $\leq$ 5 frames at 25 FPS). We source these clips from three public benchmarks: HMDB51 \cite{kuehne2011hmdb} (human actions), CelebV-HQ \cite{zhu2022celebv} (facial micro-expressions and gestures), and MM-AU \cite{fang2024abductive} (atomic understanding of momentary events).

For each video clip, we employ GPT-4 \cite{achiam2023gpt} and GPT-4o \cite{openai2024gpt4o} to synthesize two types of linguistic annotations: (i) a descriptive caption detailing salient visual dynamics, and (ii) a set of temporal QA pairs targeting transient semantics (\eg \textit{“At what timestamp does the subject snap their fingers?”, “How many times does the light flash?”, or “Does the person blink before turning their head?”}).

We apply strict quality control: all LLM-generated captions and questions are filtered by a rule-based validator that enforces grammatical correctness, temporal specificity, and grounding to visible content (\eg rejecting questions about off-screen events).

This data serves as the foundation for both supervised fine-tuning (Section \ref{sft}) and comparable reinforcement learning (Section \ref{rl}), ensuring the model receives explicit supervision on momentary visual phenomena that are often overlooked in conventional video-language corpora.


\section{Experiments}
\subsection{Implementation Details}
We adopt Qwen2.5-VL-7B and Qwen2.5-VL-3B as backbone models and leverage the built-in qwen-vl-util module of Qwen2.5-VL to process and encode visual inputs. Additionally, in constructing videos with missing key information, we employ BERT \cite{devlin2019bert} as the text encoder and utilize BLIP2 to compute the semantic similarity between video frames and their corresponding textual descriptions.

All training procedures are conducted on 8 NVIDIA A100 GPUs. In the supervised fine-tuning (SFT) stage, we set the SFT learning rate to $2 \times 10^{-4}$, the contrastive learning rate to $1 \times 10^{-4}$, and the contrastive weight to 0.1, with a total batch size of 16 ($2\times 8$). This stage requires approximately 50 GPU hours. In the reinforcement learning stage, we reduce the learning rate to $2 \times 10^{-6}$ and maintain a batch size of 8, which consumes approximately 600 GPU hours.

\subsection{Fine-grained Motion Comprehension}

\begin{table*}[t]
\centering
\small
\caption{Evaluation results on the Motionbench. Abbreviations: MR (Motion Recognition), LM (Location-related Motion), CM (Camera Motion), MO (Motion-related Objects), AO (Action Order), RC (Repetition Count). We randomly split MotionBench into ``dev" and ``test". }
\resizebox{\textwidth}{!}{%
\setlength{\tabcolsep}{3mm}{
\label{tab:motionbench}
\begin{tabular}{ccccccccccc}
\hline
Model          & LLM                  & \multicolumn{1}{c|}{\# Frames} & \multicolumn{1}{c|}{\begin{tabular}[c]{@{}c@{}}Dev AVG \\ (4020)\end{tabular}} & \begin{tabular}[c]{@{}c@{}}Test AVG \\ (4034)\end{tabular} & MR   & LM   & CM   & MO   & AO   & RC   \\ \hline
\color{dt}Random         & \color{dt}-        & \multicolumn{1}{c|}{\color{dt}-}    & \multicolumn{1}{c|}{\color{dt}0.25}         & \color{dt}0.25         & \color{dt}0.25 & \color{dt}0.25 & \color{dt}0.25 & \color{dt}0.25 & \color{dt}0.25 & \color{dt}0.25 \\ \hline
\multicolumn{11}{c}{\it{LLM: \textbf{Text} as Input}}                                                                                                                                                                                             \\ \hline
GPT-4o \cite{openai2024gpt4o}         & -               & \multicolumn{1}{c|}{-}         & \multicolumn{1}{c|}{0.33}                                                      & 0.33                                                       & 0.31 & 0.34 & 0.36 & 0.37 & 0.42 & 0.23 \\ \hline
\multicolumn{11}{c}{\it{Video VLMs : \textbf{Text + Multiple Frames} as Input}}                                                                                                                                                                                              \\ \hline
Gemini 1.5 Pro \cite{team2024gemini} & -               & \multicolumn{1}{c|}{1fps}      & \multicolumn{1}{c|}{0.51}                                                      & 0.50                                                        & 0.51 & 0.52 & 0.54 & 0.67 & 0.40 & 0.22 \\
InternVL-40B \cite{chen2024internvl} & NH-2-Yi-34B \cite{NH2Yi34B} & \multicolumn{1}{c|}{8}         & \multicolumn{1}{c|}{0.55}                                                      & 0.54                                                       & 0.54 & 0.58 & 0.49 & 0.76 & 0.41 & 0.30  \\
PLLaVA-34B \cite{xu2024pllava} & Yi-34B \cite{NH2Yi34B}    & \multicolumn{1}{c|}{16}        & \multicolumn{1}{c|}{0.52}                                                      & 0.51                                                       & 0.55 & 0.51 & 0.47 & 0.66 & 0.38 & 0.31 \\
CogVLM2-Video \cite{hong2024cogvlm2}  & LLaMA3-8B \cite{llama3modelcard} & \multicolumn{1}{c|}{24}        & \multicolumn{1}{c|}{0.41}                                                      & 0.44                                                       & 0.43 & 0.39 & 0.38 & 0.64 & 0.37 & 0.33 \\
GLM-4V-plus \cite{glm2024chatglm} & GLM4 \cite{glm2024chatglm}                 & \multicolumn{1}{c|}{30}        & \multicolumn{1}{c|}{0.54}                                                      & 0.55                                                       & 0.57 & 0.57 & 0.54 & 0.69 & 0.40  & 0.37 \\
LLaVA-NeXT \cite{zhang2024llava} & Yi-34B \cite{NH2Yi34B}  & \multicolumn{1}{c|}{32}        & \multicolumn{1}{c|}{0.48}                                                      & 0.40                                                        & 0.53 & 0.45 & 0.36 & 0.66 & 0.39 & 0.23  \\
MiniCPM-V2.6 \cite{yao2024minicpm}  & Qwen2 \cite{qwen2}   & \multicolumn{1}{c|}{64}        & \multicolumn{1}{c|}{0.52}                                                      & 0.53                                                       & 0.56 & 0.49 & 0.45 & 0.72 & 0.39 & 0.33 \\
Oryx-34B \cite{liu2024oryx}       & Yi-34B \cite{NH2Yi34B} & \multicolumn{1}{c|}{64}        & \multicolumn{1}{c|}{0.49}                                                      & 0.49                                                       & 0.48 & 0.52 & 0.44 & 0.65 & 0.42 & 0.32 \\

GLM-TE \cite{hong2025motionbench} & GLM4-9B~\cite{glm2024chatglm}              & \multicolumn{1}{c|}{16} & \multicolumn{1}{c|}{0.58}                                                      & 0.58                                                       & 0.64 & 0.59 & 0.51 & 0.69 & 0.41 & 0.39 \\
Qwen2.5-VL-3B \cite{Qwen2.5-VL}    & Qwen2.5 \cite{qwen2.5} & \multicolumn{1}{c|}{1fps}      & \multicolumn{1}{c|}{0.51}                                                      & 0.51                                                       & 0.49 & 0.51 & 0.53 & 0.44 & 0.39 & 0.36 \\
Qwen2.5-VL-7B \cite{Qwen2.5-VL}     & Qwen2.5 \cite{qwen2.5} & \multicolumn{1}{c|}{1fps}      & \multicolumn{1}{c|}{0.54}                                                      & 0.56                                                       & 0.53 & 0.55 & 0.51 & 0.69 & 0.41 & 0.35 \\
\hline

\textbf{\name-3B}  & Qwen2.5~\cite{qwen2.5}              & \multicolumn{1}{c|}{1fps} & \multicolumn{1}{c|}{0.64}                                                      & 0.64                                                       & 0.59 & 0.61 & 0.63 & 0.55 & 0.56 & 0.49 \\

\textbf{\name-7B}  & Qwen2.5~\cite{qwen2.5}              & \multicolumn{1}{c|}{1fps} & \multicolumn{1}{c|}{\textbf{0.69}}                                                      & \textbf{0.69}                                                       & \textbf{0.66} & \textbf{0.64} & \textbf{0.67} & \textbf{0.77} & \textbf{0.61} & \textbf{0.61} \\
\bottomrule
\bottomrule
\end{tabular}
}}
\end{table*}

To quantitatively assess the fine-grained action understanding capability of our proposed \name, we conduct evaluations on MotionBench \cite{hong2025motionbench}, with results summarized in Table \ref{tab:motionbench}.

MotionBench is an established benchmark specifically designed for fine-grained video action understanding. It comprises 4K open-source test samples and an additional 4K closed-source samples that require submitting model checkpoints to an official online evaluation platform for scoring. The benchmark comprehensively probes a model’s fine-grained action understanding across six distinct subtasks, including action recognition and repetition count. Evaluation on this benchmark thus provides a rigorous and multifaceted assessment of \name’s capabilities. Note that when writing this work, we did not conduct online testing, but directly made the Test metric equal to the Dev metric.

Our results demonstrate that \name~ achieves state-of-the-art performance in fine-grained action understanding. Notably, the \name-7B outperforms Qwen2.5-VL-7B \cite{Qwen2.5-VL} across all six subtasks. The most significant gain is observed in repetition counting, where \name-7B attains a score of 0.61, surpassing Qwen2.5-VL-7B \cite{Qwen2.5-VL} by 0.26. To the best of our knowledge, this marks the first instance in which any model exceeds a score of 0.6 on the repetition count subtask of MotionBench.

\subsection{Transient Events Perception}
\begin{table*}[t]
\centering
\caption{Evaluation results on the VRU-Accident benchmark. WL: Weather\&Light, TE: Traffic Environment, RC: Road Configuration, AT: Accident Type, AC: Accident Cause, and PM: Prevention Measure. The metrics highlighted in \textcolor{red}{red} are our key comparison objects.}
\label{tab:VRU-VQA}
\begin{tabularx}{\textwidth}{>{\hsize=1.5\hsize}l|*{7}{>{\centering\arraybackslash}X}}
\hline
Model (Param.) & \textbf{$Acc_{WL}$ } & \textbf{$Acc_{TE}$ } & \textbf{$Acc_{RC}$ } & \textcolor{red}{\textbf{$Acc_{AT}$ }} & \textcolor{red}{\textbf{$Acc_{AC}$ }} & \textcolor{red}{\textbf{$Acc_{PM}$ }} & \textbf{$Acc_{AVG.}$} \\
\hline\hline
LLaVA-OneVision(0.5B) \cite{li2024llava} & {72.1} & 79.9 & 40.1 & {35.9} & 34.8 & 40.8 & 50.6 \\

InternVL2.5(1B) \cite{chen2024internvl} & 69.2 & 67.7 & 39.0 & 45.8 & {21.3} & 51.0 & 49.0 \\

InternVL3(2B) \cite{zhu2025internvl3} & 70.7 & 78.3 & 53.5 & 41.2 & 50.4 & 53.2 & 57.9 \\

Qwen2.5-VL(3B) \cite{Qwen2.5-VL} & 63.8 & 72.1 & 39.4 & 47.3 & 25.6 & 44.7 & {48.8} \\

LLaVA-NeXT-Video(7B) \cite{zhang2024llava} & 71.6 & \underline{\textbf{{84.1}}}& 54.0 & 38.8 & 27.7 & {33.1} & 51.6 \\

Qwen2.5-VL(7B) \cite{Qwen2.5-VL} & 70.3 & {59.6}  & {34.1} & 60.1 & 41.4 & 46.8 & 52.1 \\

InternVL3(8B) \cite{zhu2025internvl3} & 70.0 & 81.8 & {67.8} & 64.8 & 43.7 & 58.4 & {64.4} \\

\hline

\rowcolor{gray!15}
GPT-4o-mini \cite{openai2024gpt4o} & {60.2} & 78.6 & 42.1 & 46.3 & 34.8 & 50.3 & 52.1 \\
\rowcolor{gray!15}
Gemini 1.5-flash \cite{team2024gemini}  & 65.7 & 78.5 & \underline{\textbf{{71.4}}} & {77.9} & {54.6} & 53.0 & {66.9} \\
\hline 

\textbf{\name(3B)} & 73.4 & 75.1 & 60.1 & 74.4 & 54.1 & 57.8 & 66.4 \\
\textbf{\name(7B)} & \underline{\textbf{76.7}} & 83.7 & 68.4 & \underline{\textbf{80.1}} & \underline{\textbf{61.2}} & \underline{\textbf{60.4}} & \underline{\textbf{73.7}} \\
\hline
\rowcolor{green!10}
\textbf{Human Expert} & \textbf{95.1} & \textbf{94.7} & \textbf{93.8} & \textbf{94.5} & \textbf{95.1} & \textbf{94.8} & \textbf{94.7}\\
\hline
\end{tabularx}
\end{table*}

\begin{table*}[htbp]
    \centering
    \caption{Quantitative comparisons on the dense caption task. We report ROUGE precision (P), recall (R), and F-measure (F), where P and R indicate the overlap of 4-grams with candidate and reference summaries, respectively, and F is their harmonic mean.}
     \resizebox{\textwidth}{!}{
    \begin{tabular}{l|c|c|c|ccc |ccc | ccc}
    \hline
\multirow{2}{*} {Model (Param.)} & \multirow{2}{*}{SPICE $\uparrow$} & \multirow{2}{*}{METEOR $\uparrow$}& \multirow{2}{*} {COMET $\uparrow$} &  \multicolumn{3}{c|}{ROUGE-1 $\uparrow$ } & \multicolumn{3}{c|}{ROUGE-2 $\uparrow$} & \multicolumn{3}{c}{ROUGE-L $\uparrow$} \\
 &  &  & &  P & R & F & P & R & F & P & R & F\\
\hline\hline
LLaVA-OneVision(0.5B) \cite{li2024llava} 
& 0.126
& 	0.224
& 0.647
&  0.388
&  0.384
&  0.380
& 0.092
& 0.091
& 0.090
&0.228
& 0.229
& 0.225
\\
 InternVL2.5(1B) \cite{chen2024internvl} 
& 0.132
 & 	0.236
 & 0.676
 &  0.409
&  0.403
 &  0.400
 & 0.089
 & 0.088
 & 0.087
& 0.212
 & 0.209
 & 0.207
\\
InternVL3(2B) \cite{zhu2025internvl3} 
& {{0.100}}
 & 	{{0.188}}
 & {{0.584}}
 &  0.429
 &  {{0.309}}
 &  {{0.345}}
 & 0.098
 & {{0.070}}
 & {{0.078}}
 & 0.249
 & {{0.177}}
 & 0.199
 \\
Qwen2.5-VL(3B) \cite{Qwen2.5-VL} 
& 0.149
 &  0.261
 & 0.682
 &  0.396
 &  0.446
 &  0.416
 &  0.106
 &  0.120
 &  0.111
 &  0.207
 &  0.234
 &  0.218
 \\
 LLaVA-NeXT-Video(7B) \cite{zhang2024llava} 
&  0.155
 &  0.259
 & 0.708
 &  0.443
 &  0.438
 &   0.437
 &  0.124
 &  0.123
 &  0.123
 &  0.248
 &  0.246
 &  0.245
 \\
 Qwen2.5-VL(7B) \cite{Qwen2.5-VL} 
&  {{{0.170}}}
 &  {{0.285}}
 & {{{0.721}}}
 &  0.430
 &  {{0.494}}
 &   {{{0.456}}}
 &  {{{0.126}}}
 &  {{{0.145}}}
 &  {{{0.134}}}
 &  0.233
 &  {{0.270}}
 &  {{{0.248}}}
 \\
 InternVL3(8B) \cite{zhu2025internvl3} 
&  0.159
 &  0.267
 & 0.694
 &  0.434
 &  0.454
 &   0.437
 &  0.113
 &  0.118
 &  0.114
 &  0.225
 &  0.234
 &  0.225
 \\
 \hline
 \rowcolor{gray!15}

\textbf{\name(3B)} & 0.164 & 0.267 & 0.707 & 0.412 & 0.451 & 0.449 & 0.121 & 0.144 & 0.129 &0.231 & 0.251 & 0.249 \\
 \rowcolor{gray!15}
\textbf{\name(7B)} & \underline{\textbf{0.204}} & \underline{\textbf{0.297}} & \underline{\textbf{0.751}} & \underline{\textbf{0.499}} & \underline{\textbf{0.521}} & \underline{\textbf{0.501}} & \underline{\textbf{0.153}} & \underline{\textbf{0.161}} &\underline{\textbf{0.163}} & \underline{\textbf{0.279}}& \underline{\textbf{0.276}} & \underline{\textbf{0.273}} \\   
\hline
    \end{tabular}
    
        \label{tab:VRU-caption}}
\end{table*}

To evaluate \name’s capability in perceiving transient events, we conducted experiments on the VRU-Accident benchmark \cite{kim2025vru}, which consists of 1K real-world traffic accident video clips, each ranging from tens of seconds to several minutes in duration. Despite the overall length of these videos, the critical moments corresponding to the actual accidents last no more than three seconds, highlighting the challenge of detecting and reasoning about highly transient visual events. The benchmark defines two evaluation protocols: VQA and Dense Caption generation. To thoroughly assess the model’s ability to capture transient events, we evaluated \name under both protocols, focusing on accident-centric queries and captions that directly probe understanding of the brief accident interval.

As shown in \tab \ref{tab:VRU-VQA}, \name achieved state-of-the-art performance on the accident-centric VQA task. When compared to Qwen2.5-VL, which employs the same base large language model,\name-3B improved the average VQA accuracy by 22.9\%, and \name-7B yielded a 17.8\% absolute gain. On the Dense Caption generation task (as show in \tab \ref{tab:VRU-caption}), \name-7B obtained the highest scores across all reported metrics. Additionally, \name-3B substantially outperformed other vision-language models of comparable scale, demonstrating consistent gains even with fewer parameters.

These results confirm that \name exhibits strong and robust performance in transient event perception, particularly in identifying and reasoning about extremely short-duration critical events within long video sequences.

\subsection{Generalization Evaluation}

\begin{table}[]
\caption{Performance of standard video-center tasks.}
\label{tab:general}
    \resizebox{\linewidth}{!}{%
    \setlength{\tabcolsep}{0.8mm}
     \renewcommand\arraystretch{1.3}
\begin{tabular}{@{}ccc|cc@{}}
\toprule
\multirow{4}{*}{Models}& \multicolumn{2}{c}{\multirow{2}{*}{Video Reasoning}} & \multicolumn{2}{c}{\multirow{2}{*}{Video General Benchmark}}                              \\
    & \multicolumn{2}{c}{}         & \multicolumn{2}{c}{}                                                                          \\ 
\cmidrule(l){2-5}                   & \multirow{2}{*}{\small VSI-Bench} & \multirow{2}{*}{\small VideoMMMU} & \multirow{2}{*}{\small MVBench} & \multirow{2}{*}{\small VideoMME (wo sub)} 
 \\ 
 & & & & \\
 \midrule \rowcolor{Gray}
GPT-4o \cite{openai2024gpt4o}                                                  & 34.0                         & 61.2                                                & -                                                & 71.9                                \\ \midrule
LLaMA-VID \cite{li2024llama}                                                  & -                          & -                                                     & 41.9                                      & -                                   \\
VideoLLaMA3 \cite{zhang2025videollama}                                               & -                          & -                                                   & 55.7                                               & 51.2                                \\
InternVL3-3B \cite{zhu2025internvl3}                                             & 32.1                       & 28.8                                              & -                                            & -                                   \\
Video-UTR-7B \cite{yu2025unhackable}                                           & -                          & -                                                  & 58.8                                       & 52.6                                \\
LLaVA-OneVision-7B \cite{li2024llava}                                   & 32.4                       & 33.8                                         & 56.7                                             & 58.2                                \\
 \midrule
 Qwem2.5-VL-3B \cite{Qwen2.5-VL} & 24.4 & 31.1 & 51.8 & 46.6 \\
 Qwen2.5-VL-7B \cite{Qwen2.5-VL} & 27.7 & 47.8  & 57.4 & 53.1 \\
 \textbf{\name-3B} &30.4 & 52.7 & 53.4 & 48.3 \\
 \textbf{\name-7B} &38.1 & 55.7 & 65.1 & 63.8 \\ 
\midrule
\bottomrule

\end{tabular}
}
\end{table}

In addition to the evaluations on the aforementioned tasks, we further assess our model on established benchmarks for general video understanding and reasoning. Specifically, for video understanding, we evaluate \name~ on MVBench \cite{li2024mvbench} and VideoMME \cite{fu2025video}. For video reasoning, we use VSIBench \cite{yang2025thinking} and VideoMMMU \cite{hu2025video}. The results are presented in \tab \ref{tab:general}.

Our proposed \name~ achieves strong performance on standard benchmarks for video understanding and reasoning, indicating no task-specific bias and demonstrating robust generalization.

\subsection{Ablation Study}

\begin{table}[]
    \centering
    \small
    \begin{tabular}{c|ccc}
    \hline
    \hline
        Mothod & Motionbench & VRU$_{ACC}$ & VRU$_{\text{METEOR}}$  \\
        \hline
        \name-7B &\underline{0.69} & \underline{73.7} &\underline{0.297} \\
        Ablation \textbf{A} &{0.53} &{67.1} &{0.288} \\
        Ablation \textbf{B} &0.55 & 70.4 &0.274 \\
        Ablation \textbf{C} & 0.51 & 68.4 & 0.279 \\
        Ablatiom \textbf{D} & 0.60 &69.6 & 0.291 \\
        \hline
        \hline

    \end{tabular}
    \caption{Evaluation results of ablation study. For more detailed information, please refer to the \textbf{appendix}.}
    \label{tab:placeholder}
\end{table}

To verify that all proposed components positively enhance the model’s fine-grained action and event perception capability, we conduct the following ablation studies on the aforementioned benchmarks:

\begin{itemize}
    \item \textbf{Ablation A} We omit the key-information-absent video construction and instead randomly select frames as key frames, thereby evaluating the efficacy of \name’s key frame identification module.
    \item \textbf{Ablation B} We exclude the contrastive learning loss and train the model during supervised fine-tuning using only the standard SFT objective, to assess the contribution of the contrastive learning module.
    \item \textbf{Ablation C} We substitute Comparative GRPO with standard GRPO to evaluate the impact of the comparative reinforcement learning mechanism.
    \item \textbf{Ablation D} We forgo our curated 80K task-specific dataset and train exclusively on publicly available data, isolating the performance gain attributable to the model architecture itself.
\end{itemize}

The experimental results demonstrate that each of our proposed components significantly enhances the \name’s performance in fine-grained action and event perception. Additional ablation details are provided in the \textbf{appendix}.
\section{Conclusion}
In this work, we introduced VideoPerceiver, a novel video multimodal large language model (VMLLM) designed to enhance fine-grained temporal perception in video understanding. Through a two-stage training framework that includes key-information-absent video construction, contrastive learning, and comparative reinforcement learning, VideoPerceiver significantly outperforms existing VMLLMs on tasks involving fine-grained action comprehension and transient event perception. Our method addresses critical limitations in current models by prioritizing task-relevant visual features and explicitly training the model to detect and reason about subtle temporal dynamics. The curated dataset and training strategies enable VideoPerceiver to achieve state-of-the-art performance on benchmarks such as MotionBench and VRU-Accident while maintaining strong generalization to standard video-language tasks. This work not only advances the capabilities of VMLLMs in fine-grained perception but also provides a new direction for future research in video-language understanding.
{
    \small
    \bibliographystyle{ieeenat_fullname}
    \bibliography{main}
}

\clearpage
\setcounter{page}{1}
\maketitlesupplementary

\section{Dataset Detail}
\begin{figure*}[t]
    \centering
    \includegraphics[width=\linewidth]{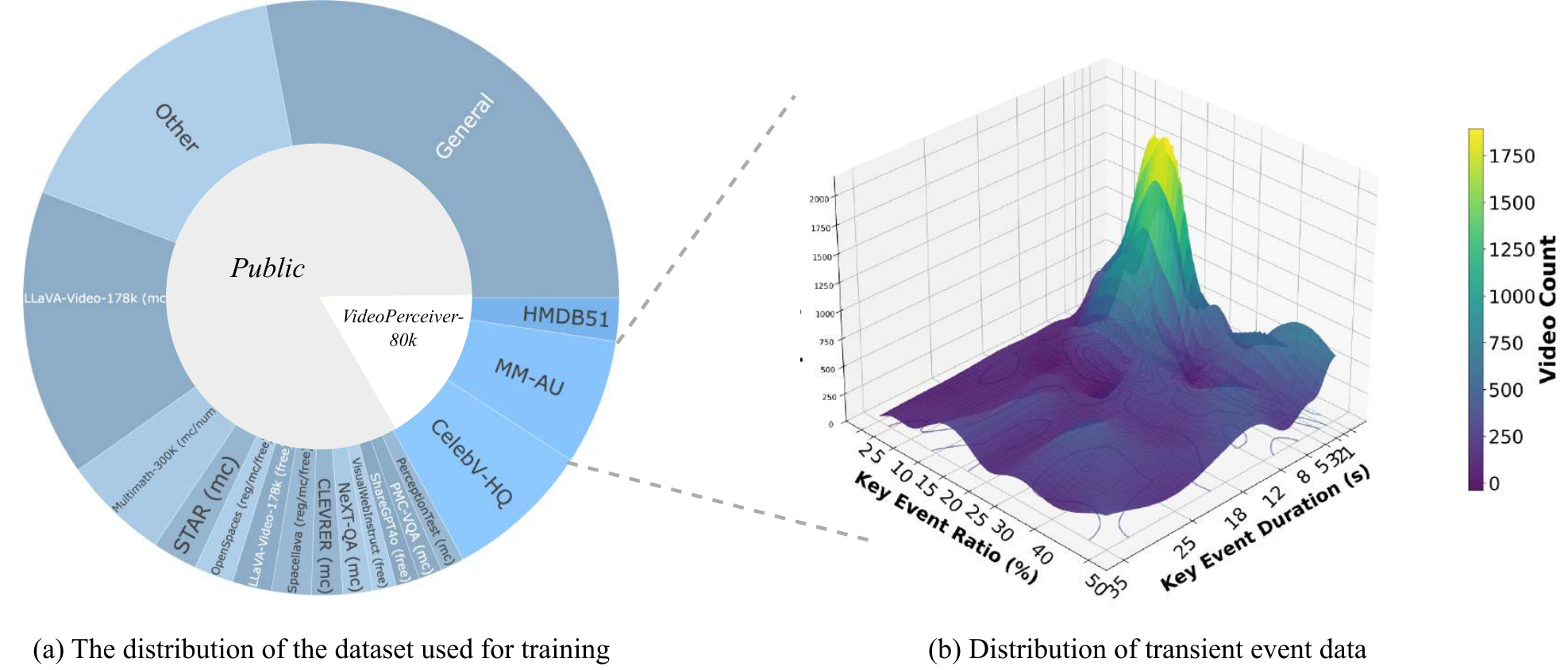}
    \caption{Summary of dataset distribution. a) The composition details of the dataset, where we will merge datasets that account for less than 1\% and replace them with the "Other" label. b) Train to understand the distribution of the MM-AU dataset using transient events. Most of the key events in the data have a low duration and account for less than 15\% of the total duration. Consistent with the understanding of transient events in real life.}
    \label{dataset_distribution}
\end{figure*}
In this section, we present the distribution of the dataset used to train \name. As shown in Figure \ref{dataset_distribution}, the dataset comprises two components: one sourced from existing open-source video datasets \cite{feng2025video}, and the other constructed by the authors using GPT-4o \cite{openai2024gpt4o} based on publicly available video data.

\section{More Ablation Study}
We additionally conducted ablation studies on two aspects: (1) the specific network block at which contrastive learning is introduced, and (2) the length of the input frame sequence. The results are summarized in Table \ref{ablation_appendix}.

In the contrastive learning ablation, the model achieves its best performance when contrastive learning is applied at the 16th block. Furthermore, whenever contrastive learning is employed across multiple blocks, inclusion of the 16th block consistently yields strong performance.

Regarding input sequence length, reducing the number of input frames (i.e., lowering temporal resolution) leads to a significant performance drop, likely because sparse sampling omits essential temporal cues. Conversely, increasing the number of input frames improves performance, suggesting that denser frame sequences provide richer motion-related information.

\begin{table*}[t]
\centering
\small
\caption{Experimental results of learning layer ablation and input frame length ablation}
\resizebox{\textwidth}{!}{%
\setlength{\tabcolsep}{3mm}{
\label{ablation_appendix}
\begin{tabular}{ccccccccccc}
\hline
Model          & CL Block Index & \multicolumn{1}{c|}{\# Frames} & \multicolumn{1}{c|}{\begin{tabular}[c]{@{}c@{}}Dev AVG \\ (4020)\end{tabular}} & \begin{tabular}[c]{@{}c@{}}Test AVG \\ (4034)\end{tabular} & MR   & LM   & CM   & MO   & AO   & RC   \\ \hline
\color{dt}Random         & \color{dt}-        & \multicolumn{1}{c|}{\color{dt}-}    & \multicolumn{1}{c|}{\color{dt}0.25}         & \color{dt}0.25         & \color{dt}0.25 & \color{dt}0.25 & \color{dt}0.25 & \color{dt}0.25 & \color{dt}0.25 & \color{dt}0.25 \\ \hline
\multicolumn{11}{c}{\it{\textbf{Comparative Learning Layer Ablation}}}                                                                                                                                                                                             \\ \hline
\name-7B         & 8               & \multicolumn{1}{c|}{1fps}         & \multicolumn{1}{c|}{0.63}                                                      & 0.63             & 0.61        & 0.59        & 0.62        & 0.72        & 0.58        & 0.58 \\ 
\textbf{\name-7B} & \textbf{16} & \multicolumn{1}{c|}{\textbf{1fps}}         & \multicolumn{1}{c|}{\textbf{0.69}} & \textbf{0.69}             & \textbf{0.66}        & \textbf{0.64}        & \textbf{0.67}        & \textbf{0.77}        & \textbf{0.61}        & \textbf{0.61}        \\
\name-7B & 24 & \multicolumn{1}{c|}{1fps} &\multicolumn{1}{c|}{0.65} & 0.65             & 0.62        & 0.63        & 0.62        & 0.75        & 0.58        & 0.59  \\
\name-7B & 15-17 & \multicolumn{1}{c|}{1fps} &\multicolumn{1}{c|}{0.67} & 0.67             & 0.64        & 0.62        & 0.65        & 0.76        & 0.60        & 0.60 \\ 
\name-7B & 14-18 &\multicolumn{1}{c|}{1fps} &\multicolumn{1}{c|}{0.68} & 0.68             & 0.65        & 0.63        & 0.66        & 0.76        & 0.60        & 0.60  \\
\name-7B &13-19 &\multicolumn{1}{c|}{1fps} &\multicolumn{1}{c|}{0.68} & 0.68             & 0.65        & 0.63        & 0.66        & 0.76        & 0.60        & 0.60 \\ 
\name-7B & 1-16 &\multicolumn{1}{c|}{1fps} &\multicolumn{1}{c|}{0.66} & 0.66             & 0.63        & 0.61        & 0.64        & 0.75        & 0.60        & 0.60 \\ 
\name-7B & 17-32 &\multicolumn{1}{c|}{1fps} &\multicolumn{1}{c|}{0.65} & 0.65             & 0.63        & 0.61        & 0.64        & 0.74        & 0.59        & 0.59 \\
\name-7B & 1-32  &\multicolumn{1}{c|}{1fps} &\multicolumn{1}{c|}{0.67} & 0.67             & 0.64        & 0.62        & 0.65        & 0.75        & 0.60        & 0.60 \\
\name-7B & 1-8, 16 &\multicolumn{1}{c|}{1fps} &\multicolumn{1}{c|}{0.65}  & 0.65             & 0.63        & 0.61        & 0.64        & 0.75        & 0.60        & 0.60 \\
\name-7B & 9-24, 16 & \multicolumn{1}{c|}{1fps} &\multicolumn{1}{c|}{0.66} & 0.66             & 0.64        & 0.62        & 0.65        & 0.75        & 0.60        & 0.60 \\
\name-7B & 25-32, 16 & \multicolumn{1}{c|}{1fps} &\multicolumn{1}{c|}{0.65} & 0.65             & 0.63        & 0.61        & 0.64        & 0.75        & 0.60        & 0.60 \\
\hline
\multicolumn{11}{c}{\it{\textbf{Frames Number Ablation}}}\\ \hline

\name-7B  & 16              & \multicolumn{1}{c|}{0.25fps} & \multicolumn{1}{c|}{0.55}                                                      & 0.55 & 0.52 & 0.50 & 0.53 & 0.65 & 0.48 & 0.49 \\ 
\name-7B & 16              & \multicolumn{1}{c|}{0.5fps} & \multicolumn{1}{c|}{0.58} & 0.58 & 0.55 & 0.53 & 0.56 & 0.68 & 0.50 & 0.51 \\
\textbf{\name-7B} & \textbf{16}              & \multicolumn{1}{c|}{\textbf{1fps}} & \multicolumn{1}{c|}{\textbf{0.69}} & \textbf{0.69} & \textbf{0.66} & \textbf{0.64} & \textbf{0.67} & \textbf{0.77} & \textbf{0.61} & \textbf{0.61} \\
\name-7B & 16              & \multicolumn{1}{c|}{2fps} & \multicolumn{1}{c|}{0.72} & 0.72 & 0.69 & 0.67 & 0.70 & 0.80 & 0.64 & 0.65 \\
\name-7B & 16              & \multicolumn{1}{c|}{4fps}& \multicolumn{1}{c|}{0.72} & 0.72 & 0.69 & 0.69 & 0.71 & 0.79 & 0.64 & 0.64 \\

\bottomrule
\bottomrule
\end{tabular}
}}
\end{table*}

\section{Large Language Models Usage Clarify}
We solemnly declare that in this article, the LLM is only used for writing assistance and text polishing (e.g., improving writing clarity and correcting grammatical errors).

\end{document}